%% file: main.tex
\documentclass{article}

\usepackage[breaklinks=true,colorlinks=true,citecolor=blue,linkcolor=blue,urlcolor=blue]{hyperref}
\usepackage[preprint]{neurips_2026}
\usepackage[utf8]{inputenc}
\usepackage[T1]{fontenc}
\usepackage{url}
\usepackage{booktabs}
\usepackage{amsmath,amsfonts,amssymb}
\usepackage{microtype}
\usepackage{xcolor}
\usepackage{graphicx}
\usepackage{multirow}
\usepackage{float}
\usepackage{array}
\usepackage{tikz}
\usetikzlibrary{arrows.meta,fit,positioning}

\title{Decode-Branch Transformers: Decoupling the Primary Prefill Path from Additional Decode Computation}

\author{%
  Liming Liu \\
  Georgia Institute of Technology
  \And
  Mingze Wang \\
  Peking University
  \And
  Tuo Zhao \\
  Georgia Institute of Technology
}

\newcommand{\decodebranch}{Decode-Branch Transformer}
\newcommand{\R}{\mathbb{R}}
\let\citepwithlink\citep
\renewcommand{\citep}[1]{\mbox{\citepwithlink{#1}}}

\begin{document}
\maketitle

\begin{abstract}
As large language models serve ever more requests, cumulative inference cost is growing relative to the one-time cost of training. In common serving regimes the two inference phases place different demands on hardware: prompt prefill runs in parallel and tends to be compute-bound, whereas autoregressive decode is sequential and memory-traffic-bound. Conventional width or depth scaling raises both costs together, since every added layer is evaluated in both phases and enlarges the weights read at every decode step. We instead ask whether additional learned computation can be allocated to continuation prediction while preserving the prompt-wide primary computation and a single persistent key--value (KV) cache. We realize this separation with the \emph{Decode-Branch Transformer}. Its primary path is a complete causal language model that alone processes the prompt and writes the KV cache; the decode branch can therefore be omitted across the prompt and activated only from its final position onward, where it adds continuation-prediction computation without writing persistent state or influencing the primary path. The paths share all major attention, MLP, and output matrices and use separate token embeddings with lightweight coupling. Grouped decode reuses each loaded weight tile and primary-cache region across both paths, so the added arithmetic does not proportionally increase the dominant memory traffic or decode latency. Across matched-token comparisons, Decode-Branch attains lower validation loss across architectures and data configurations. In MoE models, its structural separation makes the primary and branch expert fan-outs independent knobs for trading prompt cost, continuation cost, and predictive quality. We study two allocation regimes: fixing prefill expert computation while increasing decode computation, and fixing decode expert computation while reallocating the expert budget between the two paths. Together, these experiments expose a prefill--decode--quality trade-off and establish a structural opportunity for phase-specific expert allocation.
\end{abstract}

\section{Introduction}

The cost of serving large language models is determined by two very different phases. During \emph{prefill}, all prompt tokens are processed in parallel. Model weights are loaded for the batched sequence, and attention over the prompt is evaluated with highly parallel matrix operations. As context length and batch size grow, prefill is therefore primarily constrained by arithmetic throughput. During autoregressive \emph{decoding}, in contrast, tokens must be generated sequentially. Every new token requires another traversal of the model weights and another read of the growing key--value (KV) cache. Because only a small amount of computation is performed for each byte moved, decoding is typically constrained by memory bandwidth rather than peak arithmetic throughput. These distinct bottlenecks are particularly consequential for long-context inference and interactive agents, where a long prompt is followed by many sequential model calls and newly appended observations. This mismatch motivates architectures that can change the amount of computation devoted to continuation prediction without proportionally changing prompt-wide computation or persistent state.

Why should additional resources be assigned to decoding at all? Prior work shows that language-model capability can improve with more test-time computation. Chain-of-thought, self-consistency, and repeated sampling spend extra compute on intermediate or alternative trajectories \citep{wei2022cot,wang2023selfconsistency,brown2024monkeys}, while adaptive methods allocate this compute according to prompt difficulty \citep{snell2025testtime}. Pause-token methods provide more direct evidence that additional hidden computation before prediction can improve performance \citep{goyal2024pause,kim2025pause}. Together, these results motivate treating generation-time computation as a capability-scaling resource.

The systems asymmetry between prefill and decode motivates a specific architectural question. Existing test-time methods commonly spend their extra budget externally, by generating longer reasoning traces, inserting delay tokens, or sampling multiple candidates. We instead ask whether a single model can satisfy four properties simultaneously: (i) a complete primary causal path that alone processes the prompt; (ii) no persistent branch state, so the KV cache is unchanged; (iii) additional learned computation applied at continuation-prediction steps; and (iv) substantial sharing of layer weights and the primary cached state between the primary computation and the additional computation. Together, these properties define a \emph{phase-decoupled} architectural objective: make prefill and decode computation independently configurable while retaining a fixed primary prompt path and a single persistent state.

We instantiate this objective with the \decodebranch. The primary path is a complete, standard causal Transformer path that does not depend on the branch state; the decode branch reads the primary state and cache but never writes persistent state and never feeds back into the primary path. The paths share all major attention, MLP, and output matrices, but use separate token embeddings and lightweight learned coupling vectors. Because the primary path is self-contained, the decode branch can be omitted at earlier prompt positions and evaluated only at continuation-prediction steps: at the final prompt position for the first continuation token, and thereafter at each generated position. The prompt-wide primary computation and persistent KV cache therefore match the conventional model, except for the branch evaluation at the final prompt position.

This sharing is particularly consequential during decode, where weight and KV traffic dominate latency. In grouped execution, a fetched weight tile and KV region serve both paths, increasing arithmetic intensity without enlarging the set of weights and cache entries read. Decode-Branch can therefore add substantial continuation computation with only a small latency increase relative to a conventional model, rather than the near-proportional latency increase caused by adding separately stored layers or width. A mixture likelihood over the two paths' next-token distributions gives the decode-branch trajectory a direct training signal. For sparse mixture-of-experts (MoE) models, \emph{router replay} has the decode branch reuse the primary path's selected experts with its own mixture weights: this keeps the referenced expert-weight set unchanged, though the selected experts are applied to both paths' states.

The Decode-Branch Transformer's separation of prefill and decode computation becomes especially useful in MoE models, where the primary path and decode branch may activate different numbers of experts. The primary fan-out determines prompt-wide expert arithmetic, while the sum of the two fan-outs determines continuation-time expert arithmetic. The pair of fan-outs therefore defines a three-way trade-off among prefill cost, decode cost, and predictive quality. A serving system can choose a point in this allocation space according to its workload: it may preserve prompt computation while spending available decode compute on more branch experts, or hold continuation-time expert arithmetic fixed while reducing the primary fan-out for prefill-heavy workloads. This provides deployment-specific flexibility without changing the shared parameter pool or single-primary-cache structure.

Our experiments evaluate Decode-Branch through controlled NanoGPT data scaling, dense LLaMA-style model-size scaling, and sparse MoE base configurations. Under matched token budgets, Decode-Branch consistently lowers validation loss, while component ablations isolate the contributions of primary-to-branch coupling and the mixture readout. We additionally include a close training-compute comparison. For MoE, we study phase-specific expert allocation in two complementary regimes: increasing decode computation while holding prefill computation fixed, and reallocating computation between the two phases while holding decode computation fixed. These experiments expose the resulting prefill--decode--quality trade-off.

Our contributions are:
\begin{itemize}
    \item \textbf{Architecture.} We introduce Decode-Branch, an asymmetric shared-KV design that structurally decouples prompt-wide primary computation from additional continuation-prediction computation. The primary path alone processes the prompt and determines the persistent KV cache, while an branch residual trajectory reads but never writes that cache and never influences the primary path. The construction combines asymmetric shared-KV attention, shared dense weights, separate embeddings, lightweight primary-to-branch coupling, and a mixture objective; router replay extends it to MoE while preserving the primary path's selected expert set across paths.
    \item \textbf{Empirical quality.} Across controlled data scaling, dense model-size scaling, and sparse base configurations, Decode-Branch consistently reduces validation loss under matched token budgets. Component ablations isolate the contributions of primary-to-branch coupling and the mixture readout, and a complementary training-compute comparison supports the benefit of allocating computation across interacting paths.
    \item \textbf{Phase-specific MoE allocation.} Unlike prior parallel-stream designs that assign the same parameterization and compute allocation to every stream, we make the primary and branch expert fan-outs independently configurable. We study complementary regimes that respectively hold prefill or decode computation fixed, exposing a trade-off among prefill cost, decode cost, and predictive quality and enabling workload-specific expert allocation.
\end{itemize}

\section{Related work}
\label{sec:related}

\paragraph{Additional computation before prediction.}
Model and data scaling increase both training and serving cost \citep{kaplan2020scaling,hoffmann2022training}, while test-time methods show that computation can instead be allocated before committing to a prediction. Chain-of-thought exposes intermediate reasoning in generated tokens \citep{wei2022cot}; self-consistency and repeated sampling evaluate multiple candidate trajectories \citep{wang2023selfconsistency,brown2024monkeys}; adaptive methods allocate test-time compute according to prompt difficulty \citep{snell2025testtime}; and pause-token methods insert learned dummy positions that give the model additional hidden computation before producing an answer \citep{goyal2024pause,kim2025pause}. Together these results support the broad premise that next-token computation is a useful scaling resource. They generally spend that resource through additional generated positions, samples, or serial latent steps, rather than changing the internal computation of one ordinary autoregressive decode step.

\paragraph{Parallel latent computation without phase decoupling.}
Parallel-stream methods move this additional computation inside the model. Parallel Scaling applies diverse transformations to an input, processes the resulting streams with a shared backbone, and aggregates their outputs \citep{chen2025parscale}. Hidden Decoding uses independently embedded streams and retains stream-specific KV as context \citep{liu2026hidden}. State-Prediction Separation interleaves input and prediction tokens and preserves prediction-token KV within a local window \citep{monea2026sps}. These methods establish that parallel latent trajectories can improve language modeling, but their additional streams also participate in prompt processing or retain stream-specific history. Consequently, prompt computation, persistent state, or both grow with the expanded computation. Their central objective is general model-internal computation scaling, rather than holding the prompt-wide path fixed while increasing continuation-side computation.

\paragraph{Phase-decoupled latent computation.}
PHD-Transformer and Parallel Loop Transformer (PLT) are the closest precedents for separating prompt-wide persistent state from additional decode-time computation. PHD repeats tokens into original and hidden-decoding copies, retains only original-token KV for long-range attention, and computes only the original path during prefill \citep{wu2025phd}. Its training sequence uses a structured attention layout over the copies, which its kernel design rearranges to make the sparse pattern device-friendly. PLT staggers loop states across successive tokens so that different logical loops execute together during decoding; it shares first-loop global KV and augments later loops with gated sliding-window state \citep{wu2025plt}. PLT is parallel across tokens but serial across the loop dimension during training, because each loop consumes a shifted state produced by the preceding loop. Decode-Branch instead keeps two same-position residual trajectories that can be stacked and evaluated in parallel with standard causal FlashAttention \citep{dao2022flashattention}. The primary trajectory never reads the decode-branch trajectory, while the auxiliary reads same-layer primary intermediates through learned layer-wise couplings. PHD and PLT read the prediction from the final copy or loop; Decode-Branch trains a mixture in which both distributions contribute directly to the predicted next token. Thus, all three decouple persistent prompt state from additional decode computation, but realize the training graph, cross-trajectory interaction, and prediction mechanism differently. PHD is closest to Decode-Branch in architectural form; we therefore include a controlled PHD-2 implementation in the NanoGPT component ablations.

In subsequent parallel work, Langford et al. introduce \emph{free pause tokens}, a closely related phase-decoupled prediction stream that reads from the primary stream without maintaining its own persistent KV state \citep{langford2026freepause}. Their study focuses primarily on reducing pretraining overhead, particularly by phasing the prediction stream into the middle or later stages of an otherwise standard pretraining run. In contrast, our work focuses on the structural allocation of computation between prefill and decoding, and more broadly studies how additional decode computation can be scaled, including through independently configurable primary and branch expert allocations in MoE models. Their finding that the prediction stream can be introduced partway through pretraining while retaining most of its benefit provides complementary evidence for the flexibility and practicality of decode-branch architectures.

\paragraph{Other prediction and state-sharing mechanisms.}
Multi-token prediction attaches heads for future tokens \citep{gloeckle2024better}, and speculative decoding uses a draft process to reduce sequential verification cost \citep{leviathan2023fast}. Decode-Branch predicts the same next token through interacting residual trajectories, so it is neither a set of future-token heads nor a draft-and-verify method. Multi-query and grouped-query attention share KV states across query heads \citep{shazeer2019fast,ainslie2023gqa}; Decode-Branch applies a related sharing principle across full residual-state flows. Router replay further aligns the selected expert set across paths \citep{shazeer2017moe,fedus2022switch}, preserving referenced expert weights while still applying those experts to both hidden states.

\section{Background}

\subsection{Autoregressive language modeling}

Given a sequence $x_{1:T}$, an autoregressive (AR) language model factorizes its probability as
\begin{equation}
p(x_{1:T})=\prod_{t=1}^{T}p(x_t\mid x_{<t})
\end{equation}
and minimizes next-token negative log-likelihood. A causal Transformer implements each conditional distribution using a stack of self-attention and position-wise MLP layers \citep{vaswani2017attention}.

Given a matrix of residual states $H^\ell\in\mathbb R^{T\times d}$ at layer $\ell$, a Transformer forms queries, keys, and values with learned projections,
\begin{equation}
Q^\ell=H^\ell W_Q^\ell,\qquad
K^\ell=H^\ell W_K^\ell,\qquad
V^\ell=H^\ell W_V^\ell.
\end{equation}
Causal self-attention allows position $t$ to read only positions at or before $t$,
\begin{equation}
A^\ell
=
\operatorname{softmax}\!\left(
\frac{Q^\ell (K^\ell)^\top}{\sqrt{d_h}}+M_{\mathrm{causal}}
\right)V^\ell,
\end{equation}
after which an output projection, residual connection, normalization, and position-wise MLP update each token representation. Repeating this block produces the final state used to predict the next-token distribution. The same layer matrices are applied at every sequence position, while the keys and values depend on the tokens already processed.

This structure gives AR inference two distinct phases. Given a prompt of length $S$, \emph{prefill} evaluates all prompt positions together. Matrix multiplications are large enough to reuse each loaded weight across many tokens, and causal attention for the whole prompt can be computed with fused kernels such as FlashAttention without materializing the full $S\times S$ score matrix in off-chip memory \citep{dao2022flashattention}. Prefill also stores the key and value vectors produced at every layer and prompt position in a persistent KV cache.

After prefill, \emph{decode} generates one token at a time. For a new position, the model computes one new query, key, and value per layer. The new query attends to all previously cached keys and values, and the new key and value are appended to the cache. KV caching avoids recomputing earlier token states, but every generated token still traverses every layer, invokes the model weights, and reads the growing attention state. Multi-query and grouped-query attention reduce this state traffic by sharing keys and values across query heads \citep{shazeer2019fast,ainslie2023gqa}, but they do not remove the sequential dependency between generated tokens or the repeated layer-weight access.

A roofline view makes the contrast between the two phases explicit. A useful lower bound on the time of a phase is
\begin{equation}
T_{\mathrm{phase}}\gtrsim\max\!\left(\frac{F_{\mathrm{phase}}}{\Pi},\ \frac{B_{\mathrm{phase}}}{\beta}\right),
\label{eq:roofline}
\end{equation}
where $F_{\mathrm{phase}}$ is arithmetic work, $B_{\mathrm{phase}}$ is off-chip traffic, $\Pi$ is effective compute throughput, and $\beta$ is effective memory bandwidth. Under Equation~\ref{eq:roofline}, prefill at long sequence length or large effective batch tends to be compute-bound, and decode at small-to-moderate batch tends to be memory-bound because each parameter and cached KV element supports little computation before the next sequential step. These contrasting regimes motivate allocating computation separately across the two phases.

\subsection{Sparse mixture-of-experts routing}

An MoE layer replaces the single dense MLP in a Transformer block with a bank of $E$ expert MLPs and a learned router \citep{shazeer2017moe,fedus2022switch}. For a token representation $h$, the router produces logits $r(h)\in\mathbb R^E$ and selects a small set of expert indices
\begin{equation}
I(h)=\operatorname{TopK}(r(h),k),\qquad k\ll E.
\end{equation}
After normalizing the selected router scores into weights $w_e(h)$, the routed output is
\begin{equation}
\operatorname{MoE}(h)
=
\sum_{e\in I(h)}w_e(h)\operatorname{Expert}_e(h).
\label{eq:moe_layer}
\end{equation}
Some architectures additionally apply one or more shared experts to every token; these are separate from the top-$k$ routed set.

This conditional execution separates total parameter capacity from per-token arithmetic. If each routed expert contains $P_e$ parameters, the layer stores approximately $EP_e$ routed-expert parameters, but one token evaluates only $k$ experts and therefore uses approximately $kP_e$ routed parameters. Increasing $E$ can enlarge model capacity without proportionally increasing the arithmetic for an individual token, provided that $k$ remains fixed. The router and any load-balancing objective are trained jointly with the experts so that different tokens can use different subsets of the parameters.

The same top-$k$ choice has different execution characteristics during prefill and decode. During prefill, many prompt tokens are available together and can be grouped by their selected experts, providing relatively large expert batches. During autoregressive decode, each request contributes only one new token at a time; expert batches can therefore be much smaller unless many requests are served together. Expert-parallel execution must also dispatch token states to the devices holding the selected experts and combine their outputs.

\section{Decoupling prefill and decode computation with Decode-Branch}
\label{sec:method}

In this section, we present Decode-Branch as a concrete realization of phase-decoupled prefill and decode computation. Decode-Branch augments a standard primary trajectory with one branch residual-state trajectory while sharing the Transformer's large matrices. Let $d$ be the model dimension and let $S_1^\ell,S_2^\ell\in\R^{T\times d}$ denote primary and branch residual states before layer $\ell$. The two paths begin from distinct embedding tables,
\begin{equation}
S_1^0 = \operatorname{RMSNorm}(E_1[x]), \qquad
S_2^0 = \operatorname{RMSNorm}(E_2[x]).
\label{eq:embeddings}
\end{equation}
The independent embeddings are the only additional vocabulary-scale parameters. All subsequent dense projections are shared.

\subsection{Asymmetric shared-KV attention}

For normalized primary states $N_1=\operatorname{RMSNorm}(S_1^\ell)$, shared projections produce
\begin{equation}
Q_1=\operatorname{RoPE}(N_1W_Q),\quad
K_1=\operatorname{RoPE}(N_1W_K),\quad
V_1=N_1W_V.
\end{equation}
The primary attention is exactly causal self-attention,
\begin{equation}
A_1=\operatorname{CausalAttn}(Q_1,K_1,V_1)W_O.
\end{equation}
The decode branch produces only a new query. With $N_2=\operatorname{RMSNorm}(S_2^\ell)$,
\begin{align}
Q_2 &= \operatorname{RoPE}(N_2W_Q),\\
\widetilde Q_2 &= Q_2 + a^\ell\odot Q_1,\\
A_2 &= \operatorname{CausalAttn}(\widetilde Q_2,K_1,V_1)W_O,
\label{eq:asymattn}
\end{align}
where $a^\ell$ is a learned vector broadcast across tokens. Both attention computations use ordinary causal semantics and require no interleaved or method-specific attention mask, so each is compatible with standard causal attention kernels. Grouped or fused query execution lets the two paths share reads of the primary keys and values. Importantly, $S_1$ never reads $S_2$; the primary path is unchanged by whether a decode branch is evaluated.

After residual addition, $\bar S_i=S_i^\ell+A_i$, the shared MLP produces intermediate states. For a generic gated or non-gated MLP, write $H_i=f_{\mathrm{up}}(\operatorname{RMSNorm}(\bar S_i))$ for the activated intermediate representation and $W_D$ for its output projection. We use
\begin{align}
M_1 &= H_1W_D,\\
\widetilde H_2 &= H_2+b^\ell\odot H_1,\\
M_2 &= \widetilde H_2W_D+c^\ell\odot M_1,\\
S_i^{\ell+1}&=\bar S_i+M_i.
\label{eq:mlp}
\end{align}
The learned coupling vectors $a^\ell$, $b^\ell$, and $c^\ell$ strengthen information transfer from the primary path to the decode branch at the attention-query, MLP-intermediate, and MLP-output levels, respectively. This interaction remains asymmetric: it enriches the branch representation without introducing any dependence of the primary path on the decode branch.

\subsection{Mixture next-token objective}

A shared output matrix maps the final states to logits $z_1,z_2$, yielding distributions $p=\operatorname{softmax}(z_1)$ and $q=\operatorname{softmax}(z_2)$. For target token $y_t$, we minimize
\begin{equation}
\mathcal L_t=-\log\!\left(\alpha_t p_t(y_t)+(1-\alpha_t)q_t(y_t)\right).
\label{eq:mixloss}
\end{equation}
For all Decode-Branch experiments, we set
\begin{equation}
\alpha_t=\operatorname{clip}_{[0.5,1]}\!\left(\sum_v p_t(v)^2\right).
\label{eq:alpha}
\end{equation}
We use $\alpha_t$ as a confidence score based on the primary distribution's concentration. Before clipping, $\sum_v p_t(v)^2$ is exactly the probability that two independent samples from the primary distribution produce the same token---its collision probability; it is high when the distribution is concentrated and low when the primary path is uncertain. We compute $\alpha_t$ directly from the primary distribution and allow gradients to propagate through this dependence. Clipping at $0.5$ guarantees that the mixture weight on the primary path is always at least one half, so the predictive distribution used at inference---and scored by the validation loss---remains primary-dominated. We use the same mixture likelihood for training and validation.

\subsection{Combining Decode-Branch with MoE: router replay}
\label{sec:moe}

To preserve the primary path's referenced expert-weight set across both paths, we introduce \emph{router replay}. Let the primary router produce logits $r_1$ and top-$k$ expert indices $I_1$. The branch router evaluates its own logits $r_2$ but gathers routing weights only at the primary indices:
\begin{equation}
I_1=\operatorname{TopK}(r_1),\qquad
w_2=\operatorname{Normalize}\bigl(\operatorname{softmax}(r_2)[I_1]\bigr).
\end{equation}
Both paths execute the same selected experts with potentially different combination weights. Router replay thus preserves a path-specific routing signal rather than copying the primary probabilities, while keeping the set of referenced expert weights unchanged across paths. Each selected expert is applied to both hidden states, so expert arithmetic increases. Grouping the two paths' inputs allows the selected expert weights to be loaded once and reused during decoding.

\subsection{Training and inference cost analysis}
\label{sec:inference}

The per-step inference semantics make the separation between prefill and decode explicit. Let the prompt be $x_{1:S}$.
\begin{enumerate}
\item \textbf{Prompt-wide prefill.} The primary path first processes positions $1,\ldots,S$ exactly as a one-flow Transformer and writes their keys and values to the cache. The complete prompt is therefore prefetched with no decode-branch computation.
\item \textbf{Continuation boundary.} After prefill finishes, decoding begins with one branch evaluation at the final prompt token. The branch state is initialized from $E_2[x_S]$, coupled to the retained primary intermediates at position $S$, and evaluated against the completed primary KV cache. The resulting mixture predicts $x_{S+1}$. This boundary step is equivalent to adding one token-level decode evaluation, rather than adding computation to prompt-wide prefill.
\item \textbf{Autoregressive continuation.} For each subsequent position $t\ge S+1$, the primary path appends one set of keys and values, both paths form queries over the shared cache, and the mixture predicts $x_{t+1}$.
\end{enumerate}

\paragraph{Prefill.}
The prompt-wide primary graph, FLOPs, and KV construction are identical to the one-flow Transformer. Online prefill only retains the final position's primary coupling intermediates $Q_1,H_1,M_1$ for the boundary step; it performs no auxiliary computation over the prompt. The subsequent boundary evaluation adds one token-equivalent auxiliary decode operation. For a normal-length context, this constant one-token overhead is negligible relative to processing the full prompt.

\paragraph{Decode.}
At each continuation position, the primary path appends one set of keys and values. The decode branch generates queries but no persistent KV entries. The two residual states are stacked for grouped matrix operations, and both attention computations read the same primary cache. Although backbone arithmetic approximately doubles, the dominant decode data footprint---model weights and persistent KV---does not. In the memory-traffic-bound decode regime, the resulting latency increase is therefore much smaller than the FLOP increase and much smaller than that of conventional scaling that enlarges the weights or cache read at every step. Section~\ref{sec:resource_accounting} makes this distinction explicit.

\paragraph{Output distribution.}
Every continuation prediction, beginning with the first continuation token at the boundary step, uses the mixture of Equation~\ref{eq:mixloss}. The reported validation loss applies this same mixture at every scored position, so it is consistent with this serving procedure; the extra computation is paid only for continuation predictions, where batching flows can improve arithmetic intensity.

\paragraph{Training cost.}
The prompt-side savings above are an inference property and do not carry over to training. Every token position serves as the boundary for predicting its successor, so both paths are evaluated across the full training sequence. The shared attention and MLP backbone is therefore applied once to each path, giving Decode-Branch approximately twice the dominant training FLOPs of the primary model. Storing both residual trajectories similarly increases the activation memory.

The only added non-embedding parameters are the per-layer coupling vectors, which account for less than $0.1\%$ of the non-embedding parameter count. Decode-Branch also introduces a second token-embedding table. This table increases parameter and optimizer-state memory but contributes only an additional input lookup; neither it nor the coupling vectors introduce the additional matrix-based computation associated with conventional width or depth scaling. The elementwise coupling and probability-mixture operations likewise have little effect on training FLOPs or activation memory. The approximately $2\times$ cost comes from applying the same shared backbone to two states. Section~\ref{sec:nano_compute_matched} reports a close shared-backbone-compute comparison.

\subsection{Resource accounting}
\label{sec:resource_accounting}

The decode traffic for one token can be decomposed conceptually as
\begin{equation}
B_{\mathrm{decode}}=B_{\mathrm{weight}}+B_{\mathrm{state}},
\label{eq:decode_traffic}
\end{equation}
where the state term is primarily KV access for conventional attention. Decode-Branch shares both terms across its two continuation-time computations.

Because the two paths share weights and KV, grouping their states lets each shared matrix and primary KV region serve both computations. The unique weight and KV bytes referenced by a decode step remain essentially those of the primary model, while each fetched byte supports both paths. Combining this fact with the memory-bound side of Equation~\ref{eq:roofline} gives the central systems consequence: doubling useful arithmetic need not double decode time. As long as grouped execution remains below the compute roof, the decode branch consumes otherwise underutilized arithmetic throughput and adds only a small latency increment. This is the key distinction from width or depth scaling, which increases the unique weights transferred at every token and therefore directly raises the dominant decode cost.

\subsection{Experiments}
\label{sec:experiments}

We evaluate this minimal realization along three axes: training-token scaling in a controlled NanoGPT setting, dense model-size scaling in a LLaMA-style architecture, and sparse LLaMA-style MoE training with and without router replay. The dense and sparse LLaMA-style experiments use a high-quality subset of FineWeb. Architectures that allocate still more computation specifically to decoding are considered separately in Section~\ref{sec:scale_decode}.

\subsubsection{NanoGPT token scaling.}
We use the optimization setting from an earlier commit of modded-NanoGPT Track~3 \citep{jordan2024modded}. All models use 12 layers, width 768, head dimension 128, and a vocabulary of 50,257 tokens. Training uses sequences of length 1,024 and a global batch size of 512 sequences. The models are trained on FineWeb, so 3,800 optimizer steps correspond to approximately 2.0B training tokens. The standard model reaches the track's target validation loss of 3.28 at this budget. A data multiplier $D\in\{1,2,3,4,5,10,20\}$ sets the number of optimizer steps to $3{,}800D$, or approximately $2D$ billion tokens, so the Transformer and Decode-Branch see identical tokens at each $D$. Dense matrix parameters use Muon \citep{jordan2024muon}, while embeddings, output heads, and vector parameters use AdamW. Excluding token embeddings, Decode-Branch shares all matrix parameters with the Transformer and adds only the per-layer coupling vectors, which account for less than $0.1\%$ of its non-embedding parameters. Other implementation details and the schedule are shared; full optimizer settings are given in Appendix~\ref{app:optimization}.

Figure~\ref{fig:nano_steps} shows the five $D\leq5$ training trajectories. Decode-Branch finishes below the standard Transformer at every budget, and the separation persists as the token budget grows.

\begin{figure}[t]
\centering
\includegraphics[width=0.88\linewidth]{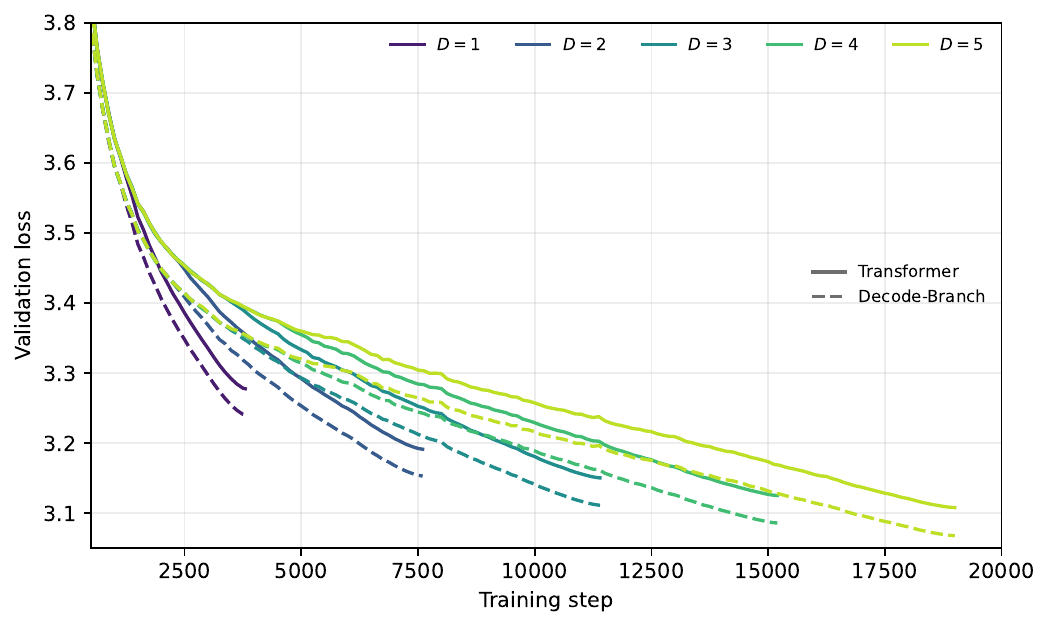}
\caption{NanoGPT validation loss versus training step at five matched token budgets in a single coordinate system. Color identifies the data multiplier $D$; solid and dashed curves denote the standard Transformer and Decode-Branch, respectively.}
\label{fig:nano_steps}
\end{figure}

To characterize the data-scaling trend, we independently fit the final points of each family to
\begin{equation}
\ell(D)=L+A D^{-\gamma},
\label{eq:scaling}
\end{equation}
which is the fixed-model-size slice of the Chinchilla scaling law: with model size held constant, its model-dependent terms are absorbed into $L$, leaving a power law in training data \citep{hoffmann2022training}. We fit the Transformer and Decode-Branch curves independently. The left panel of Figure~\ref{fig:nano_scaling_compute} fits all seven measured budgets, $D\in\{1,2,3,4,5,10,20\}$, and extends the fitted curves to $D=40$. The fitted asymptote decreases from $2.9416$ for the Transformer to $2.9013$ for Decode-Branch.

\subsubsection{Approximate training-compute match.}
\label{sec:nano_compute_matched}
During training, both paths are evaluated at every position, so one Decode-Branch step uses approximately twice the shared-backbone training FLOPs of a standard Transformer step. The scaling results above also indicate a lower fitted loss floor for Decode-Branch. As the data budget grows and the Transformer approaches its fitted floor, allocating the same training computation across two interacting paths can become more effective than spending it on additional tokens for a single flow.

\begin{figure}[t]
\centering
\begin{minipage}[t]{0.49\linewidth}
\centering
\includegraphics[width=\linewidth]{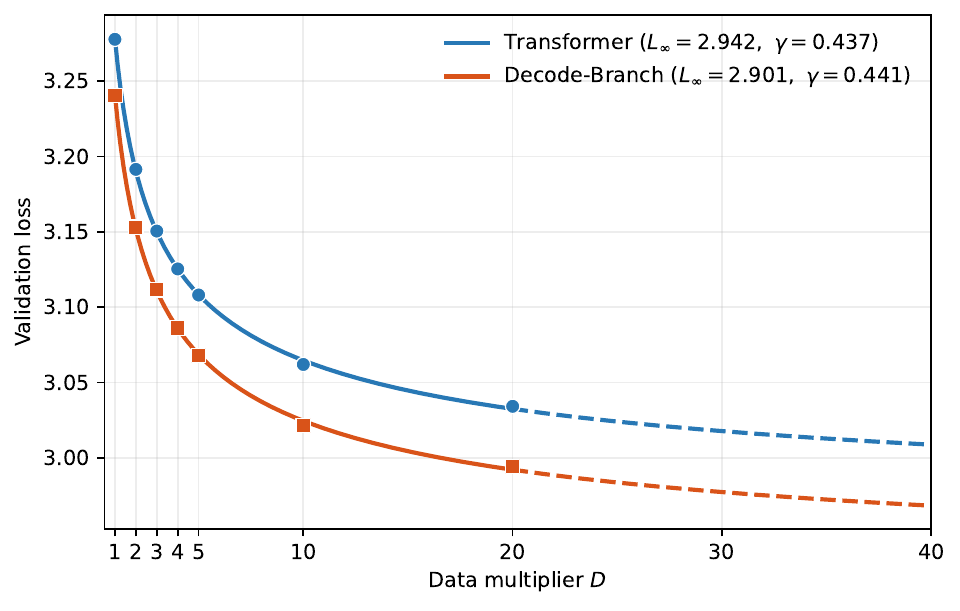}
{\small\textbf{(a)} Data scaling and fitted curves.\par}
\end{minipage}
\hfill
\begin{minipage}[t]{0.49\linewidth}
\centering
\includegraphics[width=\linewidth]{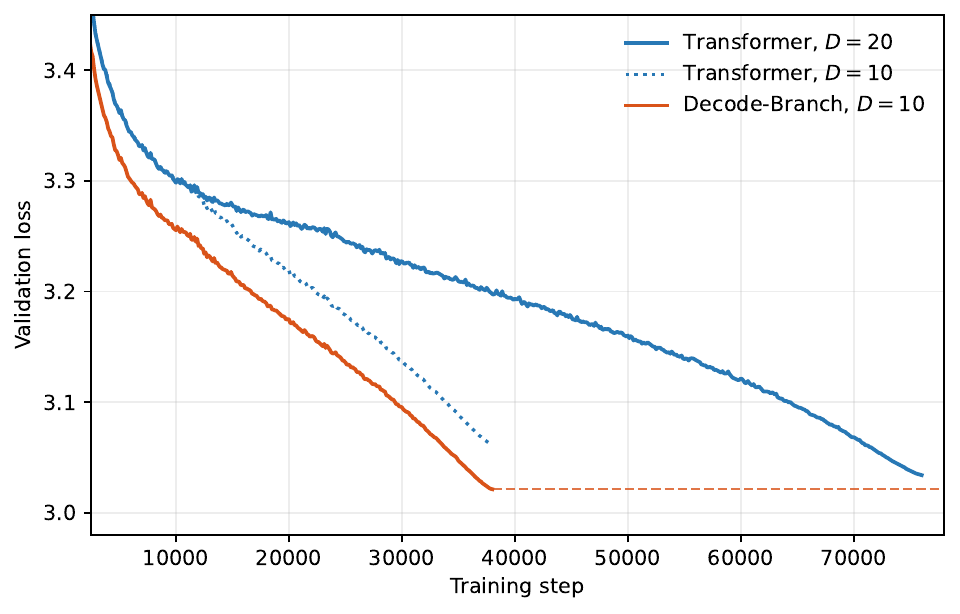}
{\small\textbf{(b)} Approximate training-compute match.\par}
\end{minipage}
\caption{NanoGPT data scaling and training-compute comparison. Left: all seven measured budgets are included in the three-parameter fits; solid lines cover the measured range and dashed lines extend the fits from $D=20$ to $D=40$. Right: the $D=10$ Transformer provides a same-step reference; the $D=20$ Transformer trains for 76,000 steps, while the $D=10$ Decode-Branch model trains for 38,000 steps with two path evaluations per step. The horizontal dashed segment extends the Decode-Branch endpoint as a reference for the $D=20$ Transformer endpoint.}
\label{fig:nano_scaling_compute}
\end{figure}

To test this high-data regime, we compare the $D=20$ Transformer with the $D=10$ Decode-Branch run. Under the approximate accounting that one Decode-Branch step performs twice the dominant shared-backbone computation of a Transformer step, their endpoints are training-compute matched. The actual training FLOPs of the $D=20$ Transformer are higher: the Decode-Branch Transformer's shared-KV branch attention computes queries without constructing a second set of keys and values. The right panel of Figure~\ref{fig:nano_scaling_compute} shows that Decode-Branch nevertheless reaches lower validation loss, supporting the benefit of allocating training computation across two interacting paths.

\subsubsection{Component ablations and PHD comparison.}
As discussed in Section~\ref{sec:related}, PHD is the prior architecture closest to Decode-Branch in form. Our PHD-2 baseline interleaves an original and a hidden-decoding copy of each token. A hidden copy can attend to the causal sequence of original copies and to itself, but not to earlier hidden copies; only the hidden copy produces the next-token logits. This interleaving expands a length-$T$ sequence to length $2T$, so the original and hidden copies of token $t$ occupy positional indices $2t$ and $2t+1$. We retain this doubled positional coordinate, rather than assigning both copies the unexpanded token position, to match the original PHD construction.

\begin{figure}[t]
\centering
\includegraphics[width=0.68\linewidth]{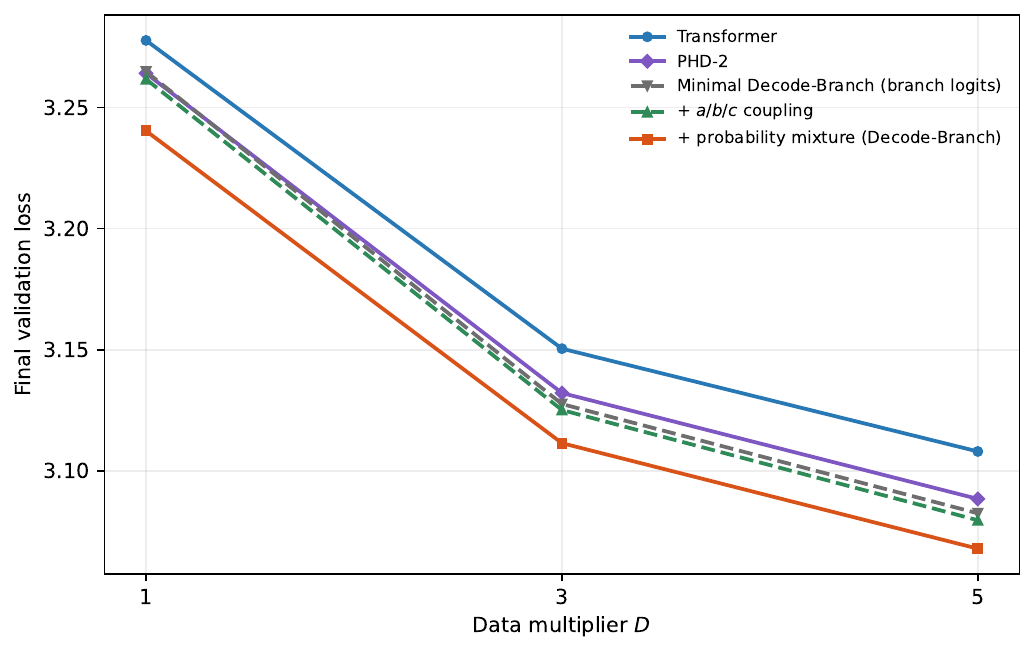}
\caption{NanoGPT component ablations at three matched token budgets, presented as an additive construction. Minimal Decode-Branch uses separate path embeddings, shared token positions, primary-to-branch shared-KV attention, and a branch-only readout. The next variant adds $a/b/c$ coupling, and the complete Decode-Branch model further adds the primary--branch probability mixture.}
\label{fig:nano_ablation}
\end{figure}

Our first ablation aligns the simplest Decode-Branch construction with PHD-2: it removes all $a/b/c$ interactions between the paths and, like PHD, predicts only from the final branch logits. The remaining differences are architectural. Decode-Branch uses separate embeddings for the two paths, assigns them the same token position rather than expanding the positional coordinate, and lets the branch query attend to the primary keys and values through primary-to-branch shared-KV attention. Figure~\ref{fig:nano_ablation} shows that this minimal Decode-Branch variant matches PHD-2 at smaller budgets and outperforms it as the data budget increases.

The second ablation adds the $a/b/c$ interactions to this branch-only model, producing a small but consistent improvement. These elementwise couplings add negligible training arithmetic and very few parameters, so we retain them in Decode-Branch. Finally, restoring the probability mixture of Equation~\ref{eq:mixloss} yields the complete model and the best performance across the tested budgets. We compare only with PHD-2 because it matches the Decode-Branch Transformer's two parallel representations and dominant training arithmetic; increasing PHD to more copies would also increase activation memory and training FLOPs, making it a different compute allocation rather than an aligned architectural comparison.

\subsubsection{Dense LLaMA model-size scaling.}
We next study model-size scaling in a LLaMA-style architecture with RMSNorm, rotary embeddings, gated MLPs, and grouped-query attention. At each labeled model size, Decode-Branch is added to the corresponding base Transformer and compared with that same base-model scale. For each base model, the training data is set to $80$ times its parameter count, keeping the data-to-model ratio fixed as scale increases. The left panel of Figure~\ref{fig:llama_model_scaling} shows consistent Decode-Branch gains over the corresponding dense baselines throughout this scaling trajectory. Full optimizer settings appear in Appendix~\ref{app:optimization}.

\subsubsection{Sparse LLaMA MoE scaling and router replay.}
\begin{figure}[t]
\centering
\begin{minipage}[t]{0.49\linewidth}
\centering
\includegraphics[width=\linewidth]{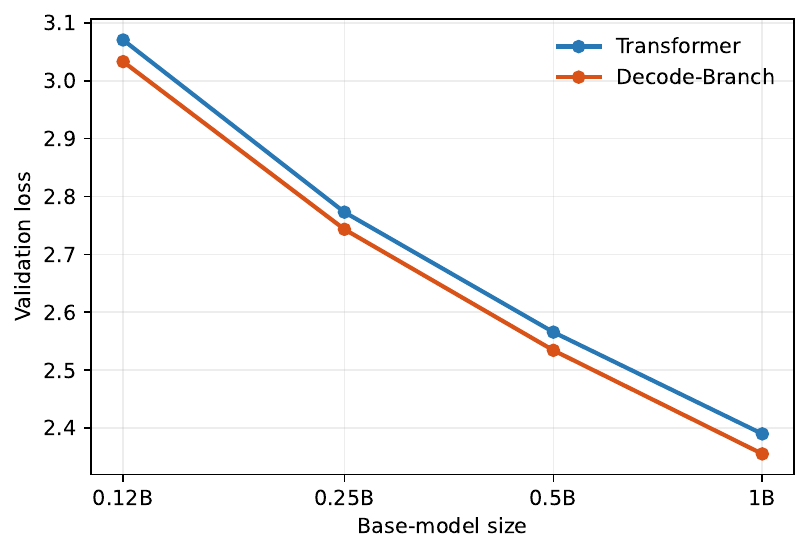}
{\small\textbf{(a)} Dense model-size scaling.\par}
\end{minipage}
\hfill
\begin{minipage}[t]{0.49\linewidth}
\centering
\includegraphics[width=\linewidth]{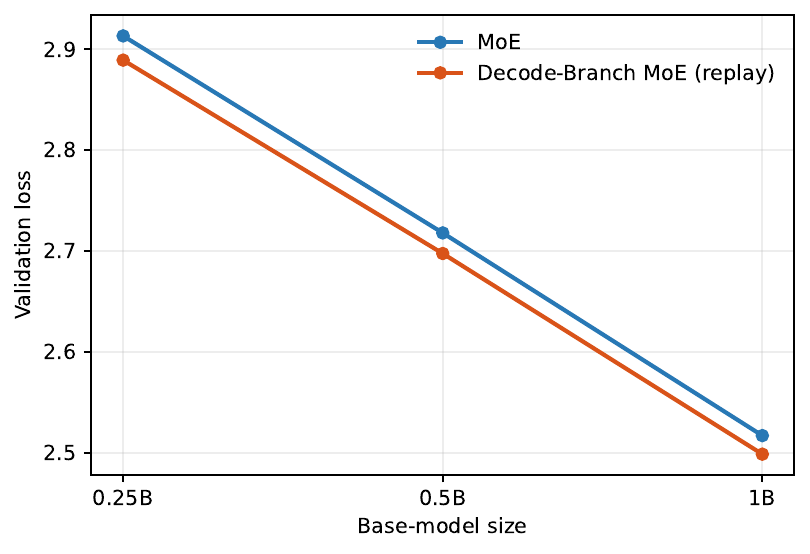}
{\small\textbf{(b)} Sparse MoE model-size scaling.\par}
\end{minipage}
\caption{LLaMA-style model-size scaling. Left: dense models trained on $80$ times the base-model parameter count. Right: sparse models trained on $40$ times the base-model parameter count, with Decode-Branch MoE using router replay. Each Decode-Branch model is compared with its corresponding baseline at the same base-model scale and token budget.}
\label{fig:llama_model_scaling}
\end{figure}

Finally, we combine Decode-Branch with a Qwen-style sparse MoE that uses routed and shared experts. Apart from the lightweight coupling vectors, the standard and Decode-Branch variants reference the same non-embedding parameter set per token. We first evaluate model-size scaling from 0.25B to 1B. At each scale, Decode-Branch MoE is added to the corresponding standard top-$4$ MoE and uses router replay, while the training data is fixed at $40$ times the base-model parameter count. The right panel of Figure~\ref{fig:llama_model_scaling} shows that the Decode-Branch MoE improvement persists across all three model sizes. Full optimizer settings appear in Appendix~\ref{app:optimization}.

\begin{figure}[t]
\centering
\includegraphics[width=\linewidth]{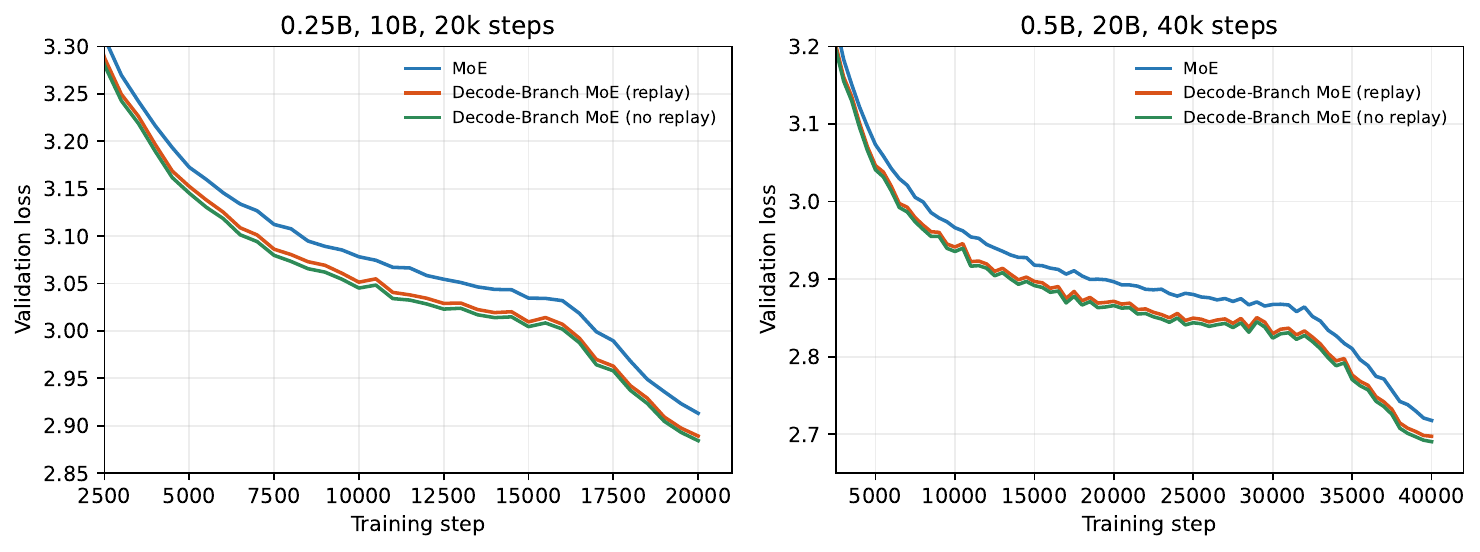}
\caption{Validation loss during sparse LLaMA-style MoE training. Both panels show completed runs for standard MoE and Decode-Branch MoE with router replay or independent branch routing (no replay).}
\label{fig:moe_training}
\end{figure}

We next examine the effect of router replay, which makes the decode branch reuse the primary path's selected experts while retaining its own mixture weights. Figure~\ref{fig:moe_training} shows that Decode-Branch MoE improves over standard MoE with either replay or independent branch routing. Independent routing provides only a modest additional gain, so replay preserves most of the improvement while keeping the referenced expert-weight set unchanged.

Across all three settings, Decode-Branch lowers matched-token validation loss in controlled NanoGPT data scaling, dense LLaMA-style model-size scaling, and sparse MoE with and without router replay, demonstrating consistent gains from the same asymmetric construction.

\section{Decoupling prefill and decode expert budgets}
\label{sec:scale_decode}

Prior parallel-stream architectures vary the representations carried by different streams---through copies, transformations, embeddings, or loop states---but treat the streams as equivalent in their allocation of model parameters and computation. Each stream is processed by the same backbone rule with the same per-stream compute budget; stream identity changes the representation, not which conditional subnetwork is activated or how much computation it receives.

Decode-Branch with MoE removes this symmetry. The paths still share one parameter pool, but path identity can determine both the subset of parameters activated and the amount of expert computation applied. Because only the primary path processes the prompt while both paths contribute during continuation, independently configuring their routers turns stream-specific computation into direct control over prefill and decode budgets. Since the two paths may now activate different numbers of experts, the allocation experiments in this section use independent routing rather than router replay.

Concretely, let the primary path activate $k_1$ routed experts and the decode branch activate $k_2$. Earlier prompt positions evaluate only the primary path, whereas the continuation boundary and each subsequent decode position evaluate both paths. Ignoring the shared expert, the routed-expert arithmetic therefore scales as
\begin{equation}
C_{\mathrm{prefill}}\propto k_1,
\qquad
C_{\mathrm{decode}}\propto k_1+k_2.
\label{eq:moe_budget}
\end{equation}
These proportionalities describe routed-expert arithmetic; shared experts, attention, and dense projections are common across allocations and are omitted.
Together, $C_{\mathrm{prefill}}$, $C_{\mathrm{decode}}$, and the resulting validation loss define a three-dimensional prefill--decode--quality trade-off surface. Different serving workloads can select different points on this surface: a prefill-constrained system may preserve prompt computation and spend more at decode, while a throughput-oriented system may hold decode computation fixed and reduce prompt cost. Rather than attempting to exhaust this full design space, we study two representative two-dimensional slices. The first fixes prefill computation while increasing decode computation; the second fixes decode computation while reallocating expert applications between the two paths.

\subsection{Fixed prefill budget: increasing decode experts}

The first regime fixes $k_1$ and varies $k_2$. Prompt-wide expert arithmetic and the primary KV cache remain unchanged, while continuation prediction receives additional expert computation. This regime is attractive when prefill cost is the binding constraint and the serving workload can spend more arithmetic at decode---for example, fragmented or lightly batched requests with available compute headroom, or single-machine deployment. It directly tests whether quality improves along
\begin{equation}
L(k_1,k_2)\quad\text{with fixed }k_1\text{ and increasing }k_2.
\end{equation}

Our sweep fixes the primary routing budget and progressively increases the independently routed auxiliary budget. Earlier prompt positions therefore retain the same expert computation, while continuation prediction receives more expert applications.

\begin{figure}[t]
\centering
\includegraphics[width=0.90\linewidth]{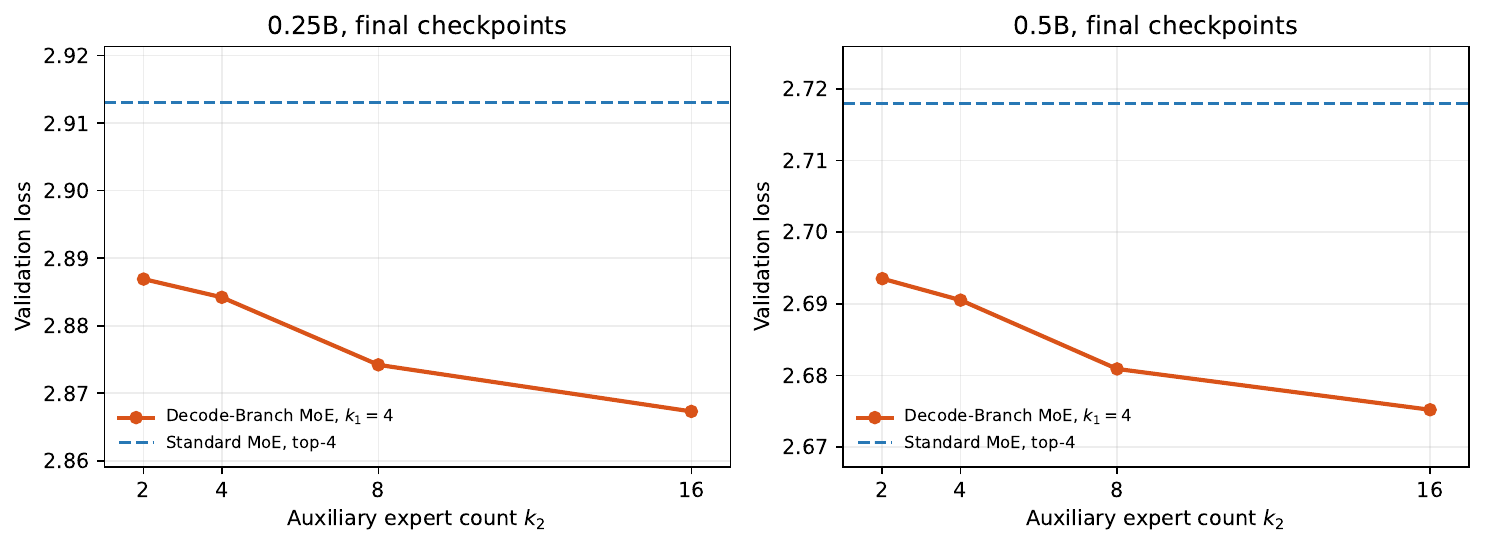}
\caption{Fixed-prefill slices at 0.25B and 0.5B, plotting final validation loss against the branch expert count. All Decode-Branch MoE runs use $k_1=4$ with independently routed $k_2\in\{2,4,8,16\}$; standard top-$4$ MoE provides the fixed-prefill baseline.}
\label{fig:fixed_prefill_allocation}
\end{figure}

Figure~\ref{fig:fixed_prefill_allocation} shows at both model scales that validation loss decreases monotonically as the branch expert budget grows. Thus, when prompt computation is fixed, additional continuation-side expert computation translates directly into better predictive quality.

\subsection{Fixed decode budget: reallocating experts}

The second regime fixes the total decode expert budget,
\begin{equation}
k_1+k_2=K,
\label{eq:fixed_decode_budget}
\end{equation}
and reallocates it between the primary path and decode branch. Every point on this slice performs the same number of routed-expert applications per decode token. Decreasing $k_1$ reduces prompt-wide expert arithmetic, while increasing $k_2=K-k_1$ preserves the decode budget. This regime targets highly batched or throughput-oriented serving, where decode capacity is fixed but reducing prompt-side expert arithmetic can increase overall efficiency.

We parameterize this slice by the normalized prefill expert fraction $k_1/K$. Relative to a standard top-$K$ MoE, this quantity is exactly the fraction of routed experts activated at each prompt position; for example, $k_1/K=1/2$ halves prefill routed-expert arithmetic while leaving the total decode allocation at $K$. This normalization also places experiments with different total budgets, such as $K=4$ and $K=8$, on a common horizontal axis. The central question is whether $L(k_1,K-k_1)$ remains stable or improves as $k_1/K$ decreases and expert budget moves from the prompt-wide primary path to the continuation-only auxiliary path. Unlike the fixed-prefill sweep, Equation~\ref{eq:fixed_decode_budget} also holds the dominant routed-expert training arithmetic constant because training evaluates both paths at every position.

\begin{figure}[t]
\centering
\includegraphics[width=0.68\linewidth]{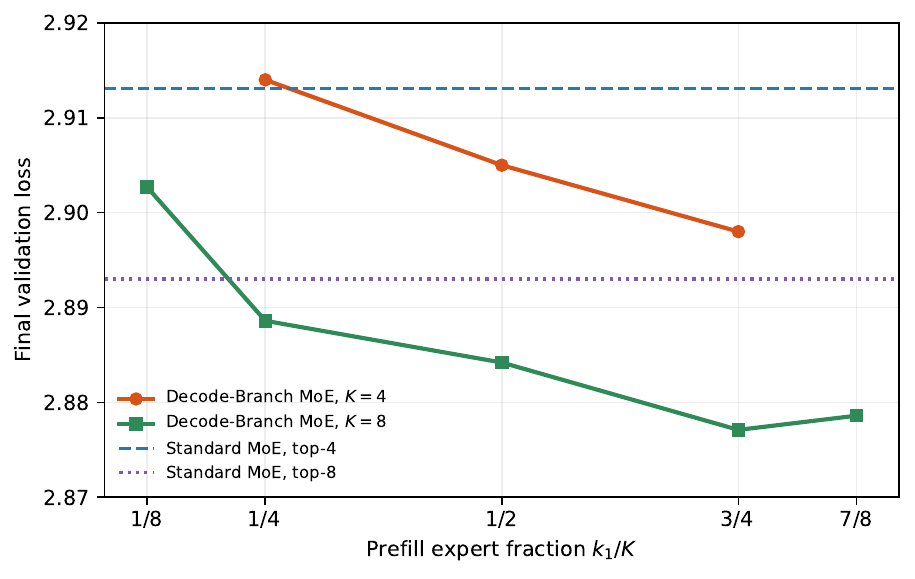}
\caption{Fixed-decode slices at 0.25B for $K=k_1+k_2\in\{4,8\}$, plotted against the normalized prefill expert fraction $k_1/K$. The $K=4$ allocations are $(1,3),(2,2),(3,1)$; the $K=8$ allocations are $(1,7),(2,6),(4,4),(6,2),(7,1)$. Standard top-$4$ and top-$8$ MoE provide the corresponding decode-expert baselines.}
\label{fig:fixed_decode_allocation}
\end{figure}

Figure~\ref{fig:fixed_decode_allocation} evaluates fixed-decode slices at two total expert budgets. The half-prefill allocation improves over the corresponding standard MoE baseline in both slices. Reducing the prefill fraction to one quarter preserves competitive quality at the smaller budget and improves over the baseline at the larger one, despite using only one quarter of the prompt-side routed-expert arithmetic. Since total decode computation is fixed, increasing $k_1$ generally strengthens the primary representation while also spending more expert computation during prefill, so quality tends to improve with the prefill fraction. This trend need not be monotonic, however, because the decode branch also contributes directly to prediction. If $k_2$ becomes too small, its contribution weakens and the model moves toward the single-flow case. The extended larger-budget slice exhibits both effects: moving away from an extremely auxiliary-heavy allocation improves substantially, performance is best near a three-quarter prefill fraction, and shifting nearly all expert computation to the primary path slightly degrades it. The optimum can therefore lie inside the allocation interval rather than at either endpoint.

These allocation sweeps target the trade-off between prompt-side computation and final loss rather than exact decode-compute matching. They retain branch attention and shared-expert evaluations because these operations reuse the primary KV and already-loaded parameters, adding only a small incremental latency in memory-bound decoding. For a strictly decode-compute-matched comparison, we introduce a Decode-Branch MoE variant in which the decode branch skips selected blocks by reusing their primary-path outputs. This variant retains strong performance; its construction and results are given in Appendix~\ref{app:compute_matched_moe}.

\section{Conclusion}

We studied how to decouple prefill and decode computation while retaining a fixed primary prompt path and a single persistent state. Decode-Branch realizes this objective with an independent primary path that alone determines prompt processing and the KV cache, and a decode branch that adds continuation-prediction computation without writing the cache or influencing the primary path. Because autoregressive decode is dominated by moving weights and KV rather than by peak arithmetic throughput, sharing both lets Decode-Branch add continuation computation with only a small incremental decode latency. Across controlled data scaling, dense LLaMA-style model-size scaling, and sparse base configurations, Decode-Branch consistently improves validation loss, while component ablations isolate the roles of primary-to-branch coupling and the mixture readout. In MoE models, the same separation enables phase-specific expert allocation. Experiments in the fixed-prefill and fixed-decode regimes expose a prefill--decode--quality frontier and show how expert computation can be selected according to the serving workload.

\section*{Acknowledgments}

We thank Damai Dai, Yuxin Fang, Defa Zhu, and Shu Zhong for insightful discussions and valuable suggestions that helped improve this work.

\bibliographystyle{plainnat}
\bibliography{ref}

\newpage

\appendix
\section{Optimization settings}
\label{app:optimization}

Within every comparison, the standard and multi-flow models use the same optimizer assignment and learning-rate schedule. Table~\ref{tab:optimization} summarizes the settings used for the reported experiments.

\begin{table}[t]
\centering
\caption{Optimizer settings. LR denotes peak or base learning rate. WSD uses linear warmup, a constant phase through 80\% of training, and linear decay to zero.}
\label{tab:optimization}
\scriptsize
\begin{tabular}{>{\raggedright\arraybackslash}p{0.16\linewidth}>{\raggedright\arraybackslash}p{0.26\linewidth}>{\raggedright\arraybackslash}p{0.22\linewidth}>{\raggedright\arraybackslash}p{0.22\linewidth}}
\toprule
Hyperparameter & NanoGPT & Dense LLaMA & LLaMA MoE \\
\midrule
Parameter assignment & Muon: hidden matrices. AdamW: embeddings, output head, and Dual coupling vectors. & Muon: matrix parameters except embeddings, LM head, and norms. AdamW: all remaining parameters. & Same as dense; expert matrices use Muon. \\
\addlinespace
AdamW LR & Embeddings: $0.3$; head: $1/320$; coupling vectors: $0.01$. & Peak $10^{-3}$ & Peak $10^{-3}$ \\
\addlinespace
Muon LR & $0.02$ & Peak $10^{-3}$ & Peak $10^{-3}$ \\
\addlinespace
Schedule & Constant for first 30\%, then linear decay to zero. & Linear warmup, then cosine decay to $5\times10^{-5}$. & WSD; linear decay to zero over final 20\%. \\
\addlinespace
Warmup steps & None & 400 / 800 / 1,000 / 1,000 at 0.12B / 0.25B / 0.5B / 1B & 400 / 800 / 1,000 at 0.25B / 0.5B / 1B \\
\addlinespace
AdamW coefficients & Embeddings/head: $\beta=(0.8,0.95)$; coupling: $(0.9,0.95)$; $\epsilon=10^{-10}$; WD $0$. & $\beta=(0.9,0.95)$; WD $0.1$ & $\beta=(0.9,0.95)$; WD $0.1$ \\
\addlinespace
Muon coefficients & Momentum $0.95$; WD $0.01$ & Momentum $0.95$; WD $0.1$ & Momentum $0.95$; WD $0.1$ \\
\bottomrule
\end{tabular}
\end{table}

\section{Additional training curves}

\subsection{NanoGPT component trajectories}

\begin{figure}[H]
\centering
\includegraphics[width=0.68\linewidth]{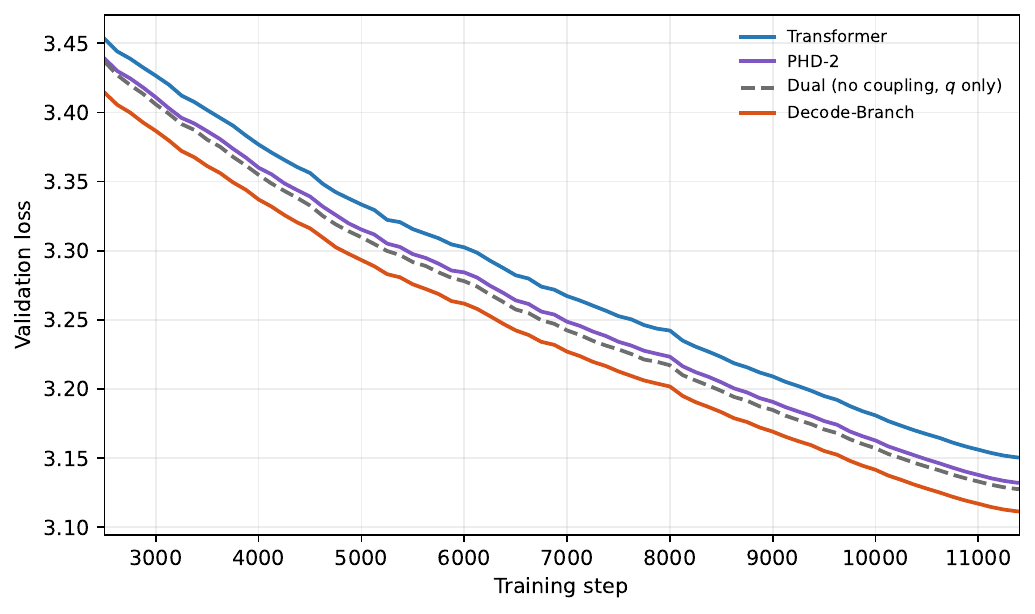}
\caption{NanoGPT validation loss versus training step at $D=3$. Curves begin at step 2,500 to emphasize the separation among the Transformer, PHD-2, minimal $q$-only Dual, and complete Decode-Branch constructions.}
\label{fig:nano_ablation_training_appendix}
\end{figure}

Figure~\ref{fig:nano_ablation_training_appendix} complements the endpoint comparison in Figure~\ref{fig:nano_ablation} with the validation-loss trajectories at $D=3$. It compares the Transformer, PHD-2, the minimal Decode-Branch construction without primary-to-branch coupling and with a $q$-only readout, and the complete Decode-Branch model.

\subsection{Fixed-prefill MoE trajectories}

\begin{figure}[H]
\centering
\includegraphics[width=0.9\linewidth]{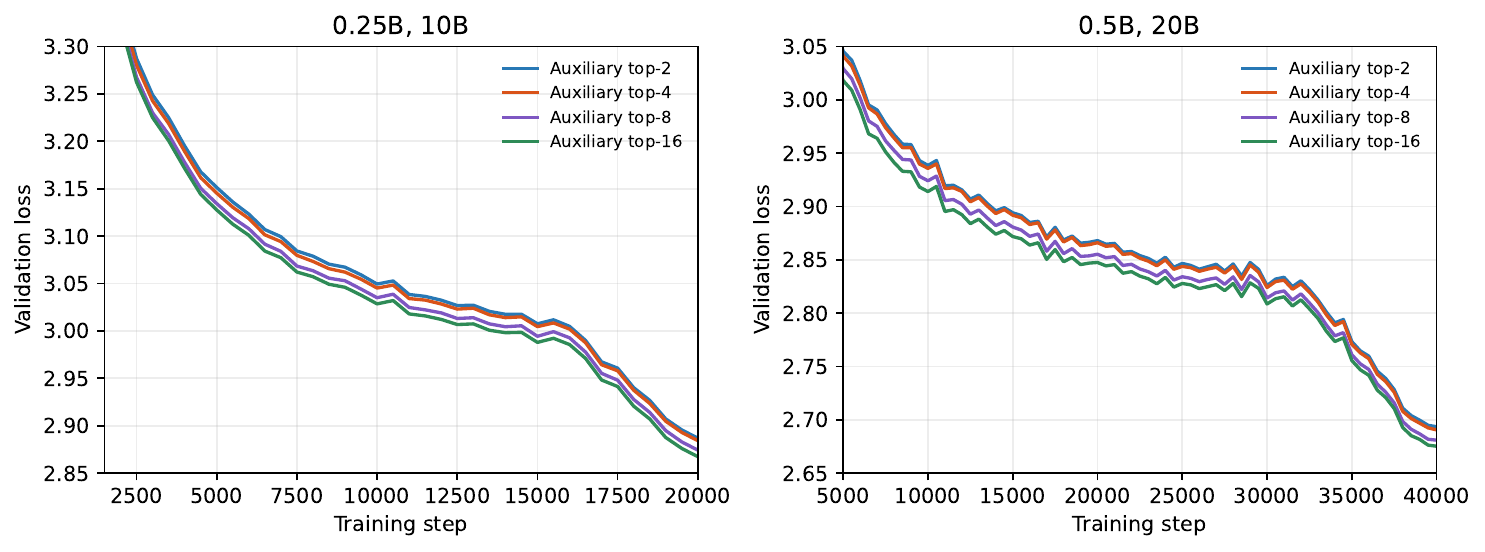}
\caption{Validation loss during the completed fixed-prefill MoE sweeps. All runs use primary top-$4$ routing and independently routed auxiliary top-$2/4/8/16$.}
\label{fig:decode_more_experts_appendix}
\end{figure}

The main text summarizes the fixed-prefill sweep using validation loss as a function of the branch expert count. Figure~\ref{fig:decode_more_experts_appendix} provides the corresponding training trajectories at both tested model scales.

\subsection{Compute-matched MoE allocation}
\label{app:compute_matched_moe}

The fixed-decode experiment in Section~\ref{sec:scale_decode} matches the total number of routed-expert applications, but its default implementation evaluates attention and the shared expert separately for both paths. This is motivated by decode execution: attention over the KV cache and the shared expert are expected to be limited primarily by memory traffic, and grouped execution can reuse their weights and cached state. We nevertheless construct a stricter control that removes their decode-branch arithmetic.

Let $N_1^\ell$ and $N_2^\ell$ be the normalized primary and branch states at layer $\ell$. In the default Decode-Branch MoE, the primary attention output $A_1^\ell$ is computed from primary queries, keys, and values, while the decode branch forms its own queries and reads the primary KV:
\begin{align}
A_1^\ell &= \operatorname{Attn}(Q_1^\ell,K_1^\ell,V_1^\ell)W_O^\ell,\\
A_2^\ell &= \operatorname{Attn}(Q_2^\ell,K_1^\ell,V_1^\ell)W_O^\ell.
\end{align}
For the compute-matched variant, we replace the second attention evaluation by a learned vector-gated copy,
\begin{equation}
A_2^\ell = g_A^\ell\odot A_1^\ell,
\label{eq:matched_attention}
\end{equation}
where $g_A^\ell\in\mathbb R^d$ is initialized to one. Thus both paths receive an attention update, but attention is evaluated only once.

We apply the same construction to the shared expert. Let $G^\ell(\cdot)$ denote the shared-expert network and let $R_i^\ell(N_i^\ell;k_i)$ denote the sum of the $k_i$ independently routed expert outputs for flow $i$. The layer's MoE outputs are
\begin{align}
M_1^\ell &= R_1^\ell(N_1^\ell;k_1)+G^\ell(N_1^\ell),\\
M_2^\ell &= R_2^\ell(N_2^\ell;k_2)
 + g_G^\ell\odot G^\ell(N_1^\ell),
\label{eq:matched_shared_expert}
\end{align}
where $g_G^\ell\in\mathbb R^d$ is another learned vector initialized to one. Equation~\ref{eq:matched_shared_expert} evaluates the shared expert only on the primary state; the auxiliary retains its own routed computation and receives a gated primary shared-expert output.

\begin{figure}[H]
\centering
\includegraphics[width=0.68\linewidth]{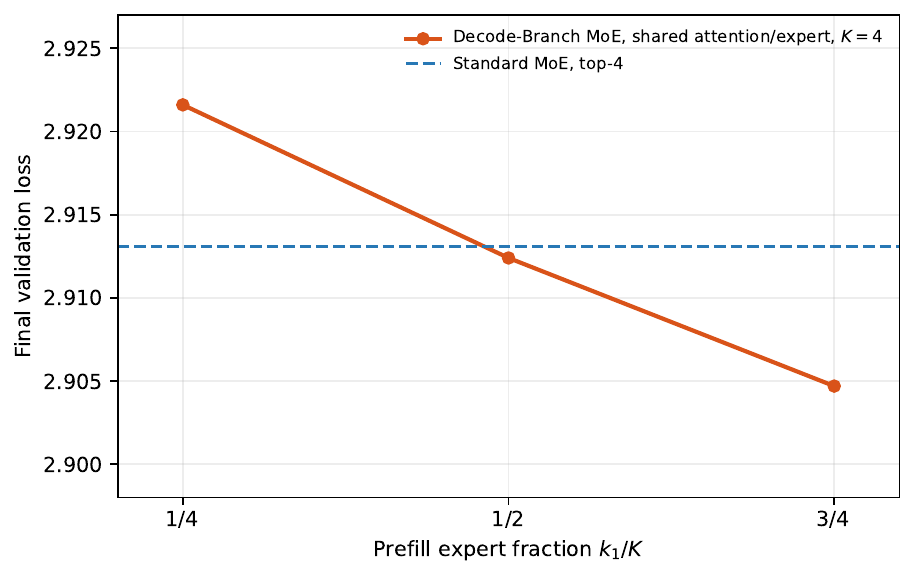}
\caption{Strictly compute-matched allocation at 0.25B with $K=k_1+k_2=4$. Decode-Branch MoE evaluates attention and the shared expert only on the primary path and sends their outputs to the decode branch through learned gates. All points match the standard top-$4$ MoE in attention, shared-expert, and routed-expert evaluations per layer; $k_1/K$ is the fraction of baseline routed-expert computation retained during prefill.}
\label{fig:compute_matched_allocation}
\end{figure}

We use the 0.25B architecture and impose
\begin{equation}
k_1+k_2=K=4.
\end{equation}
Consequently, for every layer and decode token, the standard top-$K$ MoE and this Dual variant both perform exactly one attention evaluation, one shared-expert evaluation, and $K$ routed-expert applications:
\begin{equation}
C_{\mathrm{decode}}^{\mathrm{Dual}}
=C_{\mathrm{attn}}+C_{\mathrm{shared}}+(k_1+k_2)C_{\mathrm{expert}}
=C_{\mathrm{decode}}^{\mathrm{MoE}}.
\label{eq:matched_decode_compute}
\end{equation}
The remaining branch-specific operations are lightweight elementwise gates and residual-state updates. During prefill, the decode branch is omitted, giving
\begin{equation}
\frac{C_{\mathrm{prefill,routed}}^{\mathrm{Dual}}}
{C_{\mathrm{prefill,routed}}^{\mathrm{MoE}}}=\frac{k_1}{K}.
\label{eq:matched_prefill_fraction}
\end{equation}
The horizontal axis in Figure~\ref{fig:compute_matched_allocation} is therefore exactly the retained fraction of baseline routed-expert prefill computation.

Figure~\ref{fig:compute_matched_allocation} shows a clear quality--prefill trade-off under this stricter accounting: validation loss improves as a larger fraction of the expert budget is assigned to the primary path. Notably, the half-prefill allocation matches the standard top-$4$ MoE baseline while using 50\% fewer routed-expert applications during prompt processing and no additional dominant backbone computation during decode.

\section{Other extensions of phase-decoupled computation}
\label{app:other_extensions}

The main text focuses on expert-budget allocation because it directly exposes the prefill--decode trade-off in contemporary MoE models. We additionally explore two extensions for adding continuation-side capacity: another decode branch and dense weights specific to the decode branch. Both preserve the primary prompt path and persistent KV cache while expanding the phase-decoupled design space.

\subsection{More decode branches}

For $K>2$, each decode branch $j\in\{2,\ldots,K\}$ has an independent embedding and same-position residual state but reads the same primary KV cache. We average the branch distributions,
\begin{equation}
\bar q_t(v)=\frac{1}{K-1}\sum_{j=2}^K q_{j,t}(v),
\end{equation}
and substitute $\bar q$ for $q$ in Equation~\ref{eq:mixloss}. This keeps the prompt-wide primary computation unchanged while increasing continuation arithmetic and temporary state approximately with $K$. With independent embeddings, however, each added branch also introduces another vocabulary-scale table and additional training computation.

\begin{figure}[t]
\centering
\includegraphics[width=0.68\linewidth]{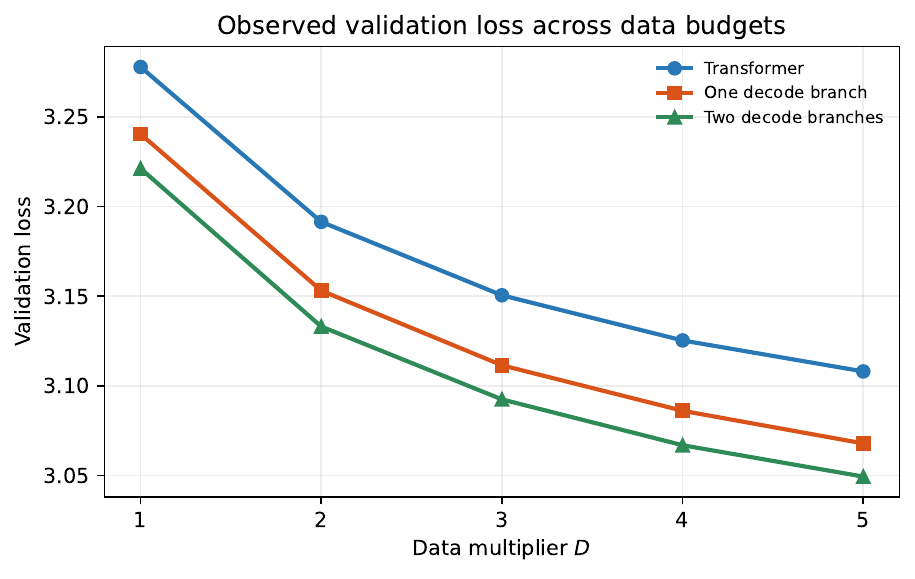}
\caption{Observed validation loss for the one- and two-branch constructions across matched training-token budgets.}
\label{fig:scaling}
\end{figure}

We instantiate two decode branches, for three residual paths in total, with coupling parameters shared across the branches. Table~\ref{tab:moreflows} and Figure~\ref{fig:scaling} show lower validation loss than the standard one-branch Decode-Branch Transformer at all five NanoGPT budgets, demonstrating that the shared-weight construction extends naturally to multiple decode branches.

\begin{table}[t]
\centering
\caption{Final validation loss for one and two decode branches at matched NanoGPT training-token budgets.}
\label{tab:moreflows}
\small
\begin{tabular}{lccccc}
\toprule
Model & $1\times$ & $2\times$ & $3\times$ & $4\times$ & $5\times$ \\
\midrule
One decode branch & 3.24035 & 3.15306 & 3.11145 & 3.08609 & 3.06789 \\
Two decode branches & \textbf{3.22114} & \textbf{3.13304} & \textbf{3.09244} & \textbf{3.06690} & \textbf{3.04940} \\
\bottomrule
\end{tabular}
\end{table}

\subsection{Auxiliary-specific dense weights}

The decode branch need not share all primary-path parameters. We give it separate query and attention-output projections and a separate MLP while retaining primary keys and values. These matrices are omitted at earlier prompt positions and evaluated at the continuation boundary and subsequent decode positions. This preserves the primary prefill graph and persistent cache, but increases deployed parameters and the unique continuation-step weight footprint.

\begin{figure}[b]
\centering
\includegraphics[width=0.68\linewidth]{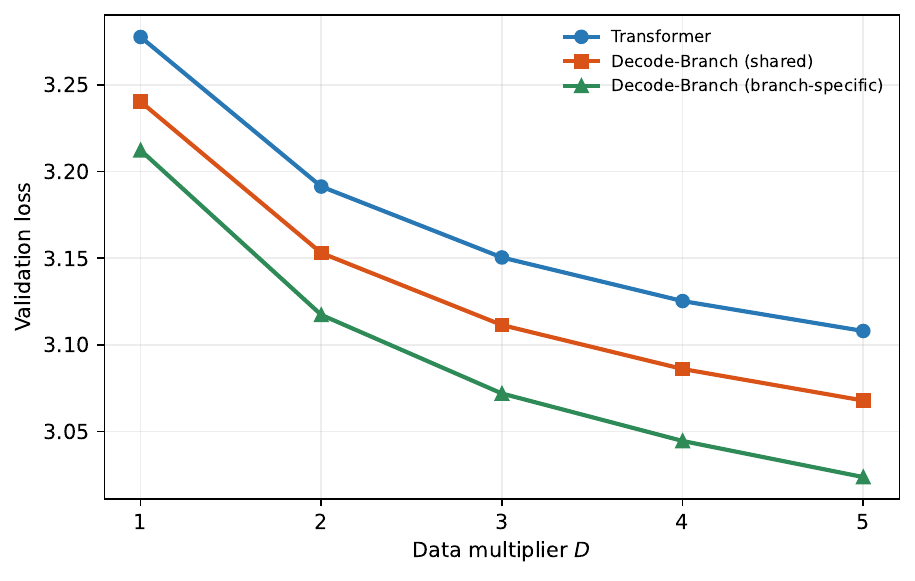}
\caption{Validation loss with shared versus branch-specific dense weights across matched NanoGPT training-token budgets.}
\label{fig:decode_dense_params}
\end{figure}

Figure~\ref{fig:decode_dense_params} shows that this variant lowers validation loss relative to shared-weight Decode-Branch throughout the measured NanoGPT range, demonstrating that branch-specific capacity provides another effective axis for continuation-side scaling.

\clearpage

\section{Connection to Hyper-Connections}
\label{app:hyper_connections}

The compute-matched construction above also admits a natural interpretation through Hyper-Connections (HC) \citep{zhu2025hyperconnections}. HC replaces one residual state by $n$ streams while retaining a single expensive block evaluation. If $X\in\mathbb R^{n\times d}$ stacks the streams, a standard HC sublayer first forms one block input $u=a^\top X$, evaluates a transformation $F(u)$, and mixes the result back into all streams. In a compact notation, its update has the form
\begin{equation}
Y=R X+bF(a^\top X),
\label{eq:standard_hc}
\end{equation}
where $a$ selects the block input, $b$ controls its write-back, and $R$ mixes the residual streams. The coefficients may be static or input-dependent. Thus, HC$\times n$ expands residual-state capacity and connectivity without evaluating attention or the MoE separately on all $n$ streams.

Decode-Branch can first be augmented with HC-style \emph{input mixing} without changing its phase-decoupled semantics. Let $X_P$ and $X_D$ denote the prefill-branch and decode-branch states before a sublayer. Rather than feeding each state directly to its transformation, form
\begin{align}
u_P &= a_{PP}^{\top}X_P, \\
u_D &= a_{DP}^{\top}X_P+a_{DD}^{\top}X_D.
\label{eq:db_input_mix}
\end{align}
The prefill input never reads $X_D$, whereas the decode input may mix both states. Setting $a_{PP}=a_{DD}=1$ and $a_{DP}=0$ recovers the original unmixed inputs, so this augmentation contains the standard Decode-Branch construction as a special parameter setting.

With input mixing, Decode-Branch becomes a masked HC$\times2$ form. For a general partition into a prefill set $P$ and a decode set $D$, write
\begin{align}
\begin{bmatrix}Y_P\\Y_D\end{bmatrix}
&=
\begin{bmatrix}R_{PP}&0\\R_{DP}&R_{DD}\end{bmatrix}
\begin{bmatrix}X_P\\X_D\end{bmatrix}
+\begin{bmatrix}b_PF_P(u_P)\\b_{DP}F_P(u_P)+b_DF_D(u_D)\end{bmatrix}.
\label{eq:masked_hc}
\end{align}
The zero upper-right block prevents decode state from entering the prefill residual streams. The prefill transformation may write to both groups through $b_P$ and $b_{DP}$, while the decode transformation writes only to $D$. Together with Equation~\ref{eq:db_input_mix}, these masks allow information to flow from prefill to decode but never in the reverse direction. For singleton $P$ and $D$, the two residual states and masked coefficients are exactly a constrained HC$\times2$ connectivity pattern; the usual Decode-Branch is obtained by choosing its original input and coupling coefficients within this family.

The two branches need not duplicate the dominant backbone computation. At the attention sublayer, the prefill branch performs the single attention evaluation and its output supplies the gated decode update in Equation~\ref{eq:matched_attention}. At the MoE sublayer, the two mixed inputs retain independent routed computations with fan-outs $k_1$ and $k_2$. The shared expert is evaluated only on the prefill input, and HC write-back can send its output directly to both streams. The strictly compute-matched variant in Equation~\ref{eq:matched_shared_expert} is recovered by retaining the shared prefill-to-decode write and masking the corresponding routed prefill-to-decode coefficient.

Figure~\ref{fig:hc2_decode_branch} makes the special-case relationship explicit. Standard HC$\times2$ forms one input for each sublayer and writes its output to two residual streams. Decode-Branch instead uses the two masked inputs above: the prefill branch remains self-contained, while the decode branch can read the prefill stream. Both branches are evaluated during continuation, whereas only the prefill branch is required over earlier prompt positions.

\begin{figure*}[t]
\centering
\resizebox{0.96\textwidth}{!}{\input{figures/hc2_decode_branch.tex}}
\caption{Decode-Branch as a masked special case of HC$\times2$. \textbf{Left:} standard HC uses one block input at each sublayer and writes the attention and MoE outputs to both residual streams. \textbf{Right:} Decode-Branch uses HC-style input mixing with a mask that keeps the prefill branch independent while allowing the decode branch to read the prefill stream. Shared and routed MoE outputs from the prefill branch, together with the decode branch's routed output, write directly to the decode residual stream. States progress alphabetically from $X$ to $Y$ to $Z$.}
\label{fig:hc2_decode_branch}
\end{figure*}

The same construction permits more than two residual streams. Figure~\ref{fig:hc4_decode_branch} illustrates an HC$\times4$ example with a $3{:}1$ partition: three streams form a self-contained prefill-branch HC group and one stream forms the decode branch. Prompt prefill instantiates only the prefill group. At the continuation boundary, the decode stream is initialized from its branch embedding and the masked prefill-to-decode connections are enabled. Attention and the shared expert remain single-evaluation operations, while routed-expert capacity is divided between the groups. Other partitions are possible; a $3{:}1$ split assigns most residual-stream capacity to the path that constructs the prompt representation while retaining one decode-only stream.

The stream ratio and the compute ratio should not be conflated. Standard HC already evaluates each expensive sublayer once regardless of $n$, so a $3{:}1$ partition is a topological allocation of residual streams, not by itself a $3{:}1$ FLOP allocation. The latter is controlled explicitly by whether attention or shared experts are replayed and by the routed fan-outs $(k_1,k_2)$. We present this correspondence as an architectural extension rather than an evaluated variant; its empirical value, initialization, and the best stream partition remain open questions.

\begin{figure*}[t]
\centering
\resizebox{0.96\textwidth}{!}{\input{figures/hc_decode_branch.tex}}
\caption{Extending the masked Decode-Branch construction to HC$\times4$. \textbf{Left:} standard HC reads all four streams at each sublayer and writes attention and MoE outputs directly to the four residual streams. \textbf{Right:} a $3{:}1$ partition assigns three streams to the self-contained prefill branch and one stream to the decode branch. Straight arrows show that all three prefill streams contribute to each decode-side input. The prefill branch's shared and routed outputs write directly to $Z_4$, alongside the decode branch's top-$k_2$ routed output. States progress alphabetically from $X$ to $Y$ to $Z$.}
\label{fig:hc4_decode_branch}
\end{figure*}

\end{document}

%% file: figures/hc2_decode_branch.tex
\begin{tikzpicture}[
  font=\sffamily\scriptsize,
  >=Latex,
  stream/.style={draw, rounded corners=2pt, minimum width=0.66cm,
    minimum height=0.46cm, inner sep=2pt},
  operation/.style={draw, rounded corners=3pt, minimum height=0.58cm,
    align=center, inner sep=3pt},
  connector/.style={draw, rounded corners=2pt, minimum height=0.48cm,
    align=center, inner sep=2pt},
  primary/.style={draw=green!48!black, fill=green!11},
  branch/.style={draw=orange!80!black, fill=orange!13},
  pflow/.style={->, line width=0.85pt, draw=green!48!black},
  dflow/.style={->, line width=0.85pt, draw=orange!82!black},
  reuse/.style={->, densely dashed, line width=0.85pt, draw=black!58}
]

\node[font=\sffamily\bfseries] at (3.05,6.75)
  {(a) Standard HC$\times2$ Transformer block};
\foreach \i/\x in {1/2.25,2/3.85}{
  \node[stream,primary] (sx\i) at (\x,0) {$X_{\i}$};
  \node[stream,primary] (sz\i) at (\x,2.65) {$Y_{\i}$};
  \node[stream,primary] (sy\i) at (\x,5.9) {$Z_{\i}$};
}
\node[connector,primary,minimum width=4.25cm] (saread) at (3.05,0.72)
  {HC attention input};
\node[operation,primary,minimum width=4.25cm] (sattn) at (3.05,1.52)
  {self-attention};
\node[connector,primary,minimum width=4.25cm] (smread) at (3.05,3.42)
  {HC MoE input};
\node[operation,primary,minimum width=1.9cm] (sshared) at (1.95,4.28)
  {shared expert};
\node[operation,primary,minimum width=1.9cm] (srouted) at (4.15,4.28)
  {routed top-$(k_1+k_2)$};
\coordinate (smerge) at (3.05,5.03);
\foreach \i in {1,2}{
  \draw[pflow] (sx\i.north) -- (saread.south);
  \draw[pflow] (sattn.north) -- (sz\i.south);
  \draw[pflow] (sz\i.north) -- (smread.south);
  \draw[pflow] (smerge) -- (sy\i.south);
}
\draw[pflow] (saread) -- (sattn);
\draw[pflow] (smread.north) -- (sshared.south);
\draw[pflow] (smread.north) -- (srouted.south);
\draw[pflow] (sshared.north) -- (smerge);
\draw[pflow] (srouted.north) -- (smerge);

\draw[black!16] (6.25,-0.48) -- (6.25,6.5);

\node[font=\sffamily\bfseries] at (11.2,6.75)
  {(b) Decode-Branch};
\node[stream,primary] (dx1) at (8.65,0) {$X_1$};
\node[stream,primary] (dz1) at (8.65,2.65) {$Y_1$};
\node[stream,primary] (dy1) at (8.65,5.9) {$Z_1$};
\node[stream,branch] (dx2) at (13.65,0) {$X_2$};
\node[stream,branch] (dz2) at (13.65,2.65) {$Y_2$};
\node[stream,branch] (dy2) at (13.65,5.9) {$Z_2$};

\node[connector,primary,minimum width=2.65cm] (paread) at (8.65,0.72)
  {HC attention input};
\node[connector,branch,minimum width=2.35cm] (daread) at (13.65,0.72)
  {HC attention input};
\node[operation,primary,minimum width=2.65cm] (pattn) at (8.65,1.52)
  {attention (evaluated once)};
\node[operation,branch,minimum width=2.35cm] (dattn) at (13.65,1.52)
  {gated attention update};

\node[connector,primary,minimum width=2.65cm] (pmread) at (8.65,3.42)
  {HC MoE input};
\node[connector,branch,minimum width=2.35cm] (dmread) at (13.65,3.42)
  {HC MoE input};
\node[operation,primary,minimum width=1.3cm] (pshared) at (7.8,4.28)
  {shared expert};
\node[operation,primary,minimum width=1.45cm] (prouted) at (9.5,4.28)
  {routed top-$k_1$};
\node[operation,branch,minimum width=2.35cm] (drouted) at (13.65,4.28)
  {routed top-$k_2$};
\coordinate (pmerge) at (8.65,5.03);

\draw[pflow] (dx1.north) -- (paread.south);
\draw[pflow] (paread) -- (pattn);
\draw[pflow] (pattn.north) -- (dz1.south);
\draw[pflow] (dz1.north) -- (pmread.south);
\draw[pflow] (pmread.north) -- (pshared.south);
\draw[pflow] (pmread.north) -- (prouted.south);
\draw[pflow] (pshared.north) -- (pmerge);
\draw[pflow] (prouted.north) -- (pmerge);
\draw[pflow] (pmerge) -- (dy1.south);

\draw[dflow] (dx1.north) -- (daread.west);
\draw[dflow] (dx2.north) -- (daread.south);
\draw[dflow] (dz1.north) -- (dmread.west);
\draw[dflow] (dz2.north) -- (dmread.south);
\draw[reuse] (pattn.east) -- node[above,text=black!58] {reuse} (dattn.west);
\draw[dflow] (daread) -- (dattn);
\draw[dflow] (dattn.north) -- (dz2.south);
\draw[dflow] (dmread) -- (drouted);
\draw[dflow] (drouted.north) -- (dy2.south);
\draw[dflow] (pshared.north east) -- (dy2.south west);
\draw[dflow] (prouted.north east) -- (dy2.south west);

\node[draw=green!35!black,rounded corners=2pt,
  fit=(dx1)(paread)(pattn)(dz1)(pmread)(pshared)(prouted)(dy1),inner sep=5pt] {};
\node[draw=orange!55!black,rounded corners=2pt,
  fit=(dx2)(daread)(dattn)(dz2)(dmread)(drouted)(dy2),inner sep=5pt] {};
\node[align=center,text=green!35!black] at (8.65,-0.58) {prefill branch};
\node[align=center,text=orange!80!black] at (13.65,-0.58) {decode branch};

\end{tikzpicture}

%% file: figures/hc_decode_branch.tex
\begin{tikzpicture}[
  font=\sffamily\scriptsize,
  >=Latex,
  stream/.style={draw, rounded corners=2pt, minimum width=0.66cm,
    minimum height=0.46cm, inner sep=2pt},
  operation/.style={draw, rounded corners=3pt, minimum height=0.58cm,
    align=center, inner sep=3pt},
  connector/.style={draw, rounded corners=2pt, minimum height=0.48cm,
    align=center, inner sep=2pt},
  primary/.style={draw=green!48!black, fill=green!11},
  branch/.style={draw=orange!80!black, fill=orange!13},
  pflow/.style={->, line width=0.85pt, draw=green!48!black},
  dflow/.style={->, line width=0.85pt, draw=orange!82!black},
  reuse/.style={->, densely dashed, line width=0.85pt, draw=black!58}
]

\node[font=\sffamily\bfseries] at (3.15,6.75)
  {(a) Standard HC$\times4$ Transformer block};
\foreach \i/\x in {1/1.35,2/2.55,3/3.75,4/4.95}{
  \node[stream,primary] (sx\i) at (\x,0) {$X_{\i}$};
  \node[stream,primary] (sz\i) at (\x,2.65) {$Y_{\i}$};
  \node[stream,primary] (sy\i) at (\x,5.9) {$Z_{\i}$};
}
\node[connector,primary,minimum width=4.7cm] (saread) at (3.15,0.72)
  {HC attention input};
\node[operation,primary,minimum width=4.7cm] (sattn) at (3.15,1.52)
  {self-attention};
\node[connector,primary,minimum width=4.7cm] (smread) at (3.15,3.42)
  {HC MoE input};
\node[operation,primary,minimum width=2.18cm] (sshared) at (1.95,4.28)
  {shared expert};
\node[operation,primary,minimum width=2.18cm] (srouted) at (4.35,4.28)
  {routed top-$(k_1+k_2)$};
\coordinate (smerge) at (3.15,5.03);
\foreach \i in {1,2,3,4}{
  \draw[pflow] (sx\i.north) -- (saread.south);
  \draw[pflow] (sattn.north) -- (sz\i.south);
  \draw[pflow] (sz\i.north) -- (smread.south);
  \draw[pflow] (smerge) -- (sy\i.south);
}
\draw[pflow] (saread) -- (sattn);
\draw[pflow] (smread.north) -- (sshared.south);
\draw[pflow] (smread.north) -- (srouted.south);
\draw[pflow] (sshared.north) -- (smerge);
\draw[pflow] (srouted.north) -- (smerge);
\draw[black!16] (6.35,-0.48) -- (6.35,6.5);

\node[font=\sffamily\bfseries] at (11.25,6.75)
  {(b) Decode-Branch HC$\times4$ with a $3{:}1$ split};
\foreach \i/\x in {1/7.65,2/8.75,3/9.85}{
  \node[stream,primary] (dx\i) at (\x,0) {$X_{\i}$};
  \node[stream,primary] (dz\i) at (\x,2.65) {$Y_{\i}$};
  \node[stream,primary] (dy\i) at (\x,5.9) {$Z_{\i}$};
}
\node[stream,branch] (dx4) at (14.45,0) {$X_4$};
\node[stream,branch] (dz4) at (14.45,2.65) {$Y_4$};
\node[stream,branch] (dy4) at (14.45,5.9) {$Z_4$};

\node[connector,primary,minimum width=3.25cm] (paread) at (8.75,0.72)
  {HC attention input};
\node[connector,branch,minimum width=2.4cm] (daread) at (14.12,0.72)
  {HC attention input};
\node[operation,primary,minimum width=3.25cm] (pattn) at (8.75,1.52)
  {attention (evaluated once)};
\node[operation,branch,minimum width=2.4cm] (dattn) at (14.12,1.52)
  {gated attention update};

\node[connector,primary,minimum width=3.25cm] (pmread) at (8.75,3.42)
  {HC MoE input};
\node[connector,branch,minimum width=2.4cm] (dmread) at (14.12,3.42)
  {HC MoE input};
\node[operation,primary,minimum width=1.45cm] (pshared) at (7.75,4.28)
  {shared expert};
\node[operation,primary,minimum width=1.65cm] (prouted) at (9.72,4.28)
  {routed top-$k_1$};
\node[operation,branch,minimum width=2.4cm] (drouted) at (14.12,4.28)
  {routed top-$k_2$};
\coordinate (pmerge) at (8.75,5.03);

\foreach \i in {1,2,3}{
  \draw[pflow] (dx\i.north) -- (paread.south);
  \draw[pflow] (pattn.north) -- (dz\i.south);
  \draw[pflow] (dz\i.north) -- (pmread.south);
  \draw[pflow] (pmerge) -- (dy\i.south);
}
\draw[pflow] (paread) -- (pattn);
\draw[pflow] (pmread.north) -- (pshared.south);
\draw[pflow] (pmread.north) -- (prouted.south);
\draw[pflow] (pshared.north) -- (pmerge);
\draw[pflow] (prouted.north) -- (pmerge);

\draw[dflow] (dx1.north) -- (daread.west);
\draw[dflow] (dx2.north) -- (daread.west);
\draw[dflow] (dx3.north) -- (daread.west);
\draw[dflow] (dz1.north) -- (dmread.west);
\draw[dflow] (dz2.north) -- (dmread.west);
\draw[dflow] (dz3.north) -- (dmread.west);
\draw[dflow] (dx4.north) -- (daread.south);
\draw[dflow] (dz4.north) -- (dmread.south);

\draw[reuse] (pattn.east) -- node[above,text=black!58] {reuse} (dattn.west);
\draw[dflow] (daread) -- (dattn);
\draw[dflow] (dattn.north) -- (dz4.south);
\draw[dflow] (dmread) -- (drouted);
\draw[dflow] (drouted.north) -- (dy4.south);
\draw[dflow] (pshared.north east) -- (dy4.south west);
\draw[dflow] (prouted.north east) -- (dy4.south west);

\node[draw=green!35!black,rounded corners=2pt,
  fit=(dx1)(dx2)(dx3)(paread)(pattn)(dz1)(dz2)(dz3)(pmread)(pshared)(prouted)(dy1)(dy2)(dy3),
  inner sep=5pt] {};
\node[draw=orange!55!black,rounded corners=2pt,
  fit=(dx4)(daread)(dattn)(dz4)(dmread)(drouted)(dy4),
  inner sep=5pt] {};
\node[align=center,text=green!35!black] at (8.75,-0.58)
  {prefill branch};
\node[align=center,text=orange!80!black] at (14.12,-0.58)
  {decode branch};

\end{tikzpicture}